\pdfoutput=1
\documentclass[runningheads]{llncs}
\usepackage{cite}
\usepackage{comment}
\usepackage{algorithm}
\usepackage{algorithmic}
\usepackage{amsmath}
\usepackage{amssymb}
\usepackage{multirow}
\usepackage{subfigure}
\usepackage{graphicx}
\usepackage{float} 

\usepackage{url}
\usepackage[misc]{ifsym}
\usepackage{hyperref}
\makeatletter
\def\UrlAlphabet{%
      \do\a\do\b\do\c\do\d\do\e\do\f\do\g\do\h\do\i\do\j%
      \do\k\do\l\do\m\do\n\do\o\do\p\do\q\do\r\do\s\do\t%
      \do\u\do\v\do\w\do\x\do\y\do\z\do\A\do\B\do\C\do\D%
      \do\E\do\F\do\G\do\H\do\I\do\J\do\K\do\L\do\M\do\N%
      \do\O\do\P\do\Q\do\R\do\S\do\T\do\U\do\V\do\W\do\X%
      \do\Y\do\Z}
\def\UrlDigits{\do\1\do\2\do\3\do\4\do\5\do\6\do\7\do\8\do\9\do\0}
\g@addto@macro{\UrlBreaks}{\UrlOrds}
\g@addto@macro{\UrlBreaks}{\UrlAlphabet}
\g@addto@macro{\UrlBreaks}{\UrlDigits}
\makeatother

\usepackage[utf8x]{inputenc}
\usepackage{color}

\begin{document}
	\title{ Smoothed Multi-View Subspace Clustering
	}
	%
	\author{
	Peng Chen\inst{1} \and
	Liang Liu\inst{2} \and
	Zhengrui Ma\inst{2} \and
	Zhao Kang\inst{2, 3{(\textrm{\Letter})}}
	}

	\authorrunning{C. Peng et al.}
	%
	\institute{
	Jangsu Automation Research Institute, Lianyungang, Jiangsu, China	\and 
	University of Electronic Science and Technology of China, Chengdu, Sichuan, China
	\and
	Trusted Cloud Computing and Big Data Key Laboratory of Sichuan Province, Chengdu, Sichuan, China\\
		\email{Zkang@uestc.edu.cn}
	}
	
	\maketitle              
	\begin{abstract}
		In recent years, multi-view subspace clustering has achieved impressive performance due to the exploitation of complementary imformation across multiple views. However, multi-view data can be very complicated and are not easy to cluster in real-world applications. Most existing methods operate on raw data and may not obtain the optimal solution. In this work, we propose a novel multi-view clustering method named smoothed multi-view subspace clustering (SMVSC) by employing a novel technique, i.e., graph filtering, to obtain a smooth representation for each view, in which similar data points have similar feature values. Specifically, it retains the graph geometric features through applying a low-pass filter. Consequently, it produces a ``clustering-friendly" representation and greatly facilitates the downstream clustering task. Extensive experiments on benchmark datasets validate the superiority of our approach. Analysis shows that graph filtering increases the separability of classes.
		
	\end{abstract}
	\section{Introduction}
	
	{A}{s} one of the most fundamental tasks in data mining, pattern recognition, and machine learning, clustering has been extensively used as a preprocessing step to facilitate other tasks or a standalone exploratory tool to reveal underlying structure of data \cite{jain2010data}. According to their intrinsic similarities, it partitions unlabeled data points into disjoint groups. Nevertheless, clustering performance can be easily affected by many factors, including data representation, feature dimension, and noise \cite{zhang2019flexible}. Clustering is still a challenging task though numerous progresses have been made in the past few decades \cite{zhou2020self,wang2019discovering}. 
	
	Specifically, there are a number of classical clustering algorithms, including K-means clustering, DBSCAN, agglomerative clustering, spectral clustering. 
    It is well-known that K-Means works best for data evenly distributed around some centroids \cite{yang2017towards,liu2020nearly}, which is hard to satisfy in real-world data. Afterwards, numerous techniques, including kernel trick, principal component analysis, and canonical correlation analysis, are applied to map the raw data to a certain space that better suits K-means. In recent years, spectral clustering has become popular due to its impressive performance and well-defined mathematical framework \cite{ng2002spectral}. Many variants of spectral clustering have been developed in the literature \cite{chen2019labin}. The performance of such methods heavily depend on the quality of similarity graph \cite{kang2020robust}. Some recent efforts are made to automatically learn a graph from data \cite{kang2018unified,ren2020simultaneous}.
	
	During the last decade, subspace clustering (SC) has attracted considerable attention due to its capability in partitioning high dimensional data \cite{lv2021pseudo}. It assumes that data lie in or near some low-dimensional subspaces and each data point can be expressed as a linear combination of others from the same subspace. The learned coefficient matrix $Z$ is treated as a similarity graph and then it is fed to a spectral clustering algorithm. Consequently, each cluster corresponds to one subspace \cite{li2015robust}. Two seminal subspace clustering models are sparse subspace clustering (SSC) \cite{elhamifar2013sparse} and low-rank representation (LRR) \cite{liu2012robust}. For $Z$, SSC enforces $\ell_1$-norm to achieve a sparse solution, while LRR applies the nuclear norm to obtain a low-rank representation. 
	To capture non-linear relationships, some kernel-based subspace clustering \cite{kang2020structured} and deep neural networks based methods have been developed \cite{kang2020relation,ji2017deep}. 
	
	In the era of big data, increasing volume of data are collected from multiple views. For instance, news can be reported in different languages and in the form of texts, images, and videos \cite{kang2019multiple,wen2020adaptive}; an image can be represented by different features, e.g., GIST, LBP, Garbor, SIFT, and HoG \cite{chen2020multi,chen2019jointly}. Therefore, many multiview subspace clustering methods have been proposed to explore the consensus and complementary information across multiple views \cite{tan2020unsupervised}. For example, \cite{gao2015multi} learns a graph for each view and let them share a unique cluster indicator matrix; \cite{cao2015diversity} explicitly incorporates the diversity of views; \cite{kang2020partition} 
	performs information fusion in partition space; \cite{zhang2018generalized} performs learning in a latent space. Nevertheless, these methods have a high time complexity. Until recently, a linear algorithm for multi-view subspace clustering was developed by Kang et al. \cite{kang2019large,kang2021structured}.
	
	We observe that most of the existing multi-view subspace clustering methods operate on the raw data. In real-world applications, complex high-dimensional data in original data space itself might not satisfy self-expression property so that the data are not be separable into subspaces. Therefore, some studies perform subspace clustering in an alternative representation space instead of the original domain. The motivation is that the data are separable after being projected into a new domain. 
	In particular, there are two categories of methods, i.e., \cite{patel2015latent} learns a tight-frame for subspace clustering; \cite{ji2017deep} learn a latent representation via auto-encoders. The former class are shallow techniques which lack high discriminative capability while the later are deep learning approaches which involve large number of parameters and are computationally expensive.
	
	In this paper, we manage to find a smooth representation for multi-view subspace clustering, in which similar samples will have similar representations. Therefore, this representation is ``clustering-friendly'', i.e., it is easy to cluster. To this end, we preserves the graph geometric features by applying a low-pass filter. Putting it differently, the structure information carried by similarity graph is employed to extract meaningful data representation for clustering. To verify the effectiveness of our approach, we examine it on multi-view model. Notably, the proposed strategy is general enough to integrate with various multi-view subspace clustering models. Extensive experiments and analysis demonstrate our superiority.
	
	The main contributions of this paper are summarized as follows:
	\begin{itemize}
		\item{A graph filtering framework for multi-view subspace clustering is developed, which provides a new representation learning strategy.}
		\item{{Remarkable improvements brought by graph filtering are demonstrated on multi-view datasets.}}
        \item{Experimental analysis shows that the graph filtering obviously pushes different clusters apart.}
	\end{itemize}
	
	\section{Preliminaries and Related Work}
	\subsection{Graph Filtering}
	Given an affinity matrix
	$W\in\mathbb{R}^{n\times n}$ of an undirected graph $G$, where $w_{ij} = w_{ji} \geq 0$ and $n$ is the number of nodes, the degree matrix and symmetrically normalized Laplacian can be derived as $D=\emph{diag}(d_1,\cdots,d_n)$ and $L_{s} = I - D^{-\frac{1}{2}}WD^{-\frac{1}{2}}$, where $d_{i} = \sum_{j=1}^{n}w_{ij}$. Since $L_{s}$ is real-symmetric, it could be eigen-decomposed as $L_{s} = U\Lambda U^{\top}$, where $U=[\emph{\textbf{u}}_1,\cdots,\emph{\textbf{u}}_n]$ is an unitary matrix and the eigenvalues $\Lambda=\emph{diag}(\lambda_1,\cdots,\lambda_n)$ are sorted in increasing order. The eigenvectors, i.e., $\{\emph{\textbf{u}}_1,\cdots,\emph{\textbf{u}}_n\}$, are the Fourier basis associated with the graph $G$ and the corresponding eigenvalues $\lambda_{i}$ indicate their frequencies \cite{shuman2013emerging}.\par
	A \emph{graph signal} is in fact a mapping function \emph{\textbf{f}} defined on the nodes, i.e., $ \emph{\textbf{f}} = [f(v_{1}),f(v_{2}),...,f(v_{n})]^{\top}$. When a feature matrix
	$X=[\emph{\textbf{x}}_1,\cdots,\emph{\textbf{x}}_n]^\top\in\mathbb{R}^{n\times m}$ is given in real-world application, each feature dimension can be considered as a signal on the graph nodes. A graph signal is then denoted as a linear combination of the Fourier basis of the graph,
	\begin{equation}
		\emph{\textbf{f}} = \sum_{i=1}^{n}c_{i}\emph{\textbf{u}}_{i} = U\emph{\textbf{c}},
	\end{equation}
	where $\emph{\textbf{c}}= [c_{1},c_2,...,c_n]^\top$ represents the Fourier coefficient and the absolute value of $c_i$ suggests the strength of the basis signal $\emph{\textbf{u}}_{i}$ in the graph signal \emph{\textbf{f}}. Thus, we can measure the smoothness of graph signal \emph{\textbf{f}} in frequency domain by
	\begin{equation}
		\begin{aligned}
			E_{f} &= \frac{1}{2}\sum_{i,j=1}^{n}w_{ij}\|\frac{{f}_{i}}{\sqrt{d_{i}}}-\frac{{f}_{j}}{\sqrt {d_{j}}}\|_2^{2} =  {\emph{\textbf{f}}}^{\top} L_{s} \emph{\textbf{f}}\\
			&= {(U\emph{\textbf{c}})}^{\top}L_{s}U\emph{\textbf{c}} = \sum_{i=1}^{n} {c_i}^{2}\lambda_{i}.
		\end{aligned}
	\end{equation}
    This formulation shows that smooth signal should have small eigenvalue. Therefore, a smooth signal $\emph{\textbf{f}}$ mainly contains low-frequency basis signals \cite{zhang2019attributed}.\par
	In practice, the natural signal is often smooth since the graph signal values change slowly between connected neighbor nodes. If we want to get a
    smooth signal after filtering, a low-pass graph filter $G$ can be applied. Define $h(\lambda_{i})$ as the low-pass frequency response function and its value should decrease when the frequency $\lambda_i$ increase. Since the eigenvalues of $L_s$ fall into $[0,2]$, a simple low-pass filter could be designed as $h(\lambda_{i}) = (1 - \frac{\lambda_{i}}{2})^{k}$, where $k>0$ is an integer capturing the $k$-hop neighborhood relationship \cite{ma2020towards}. Then, the filtered $\bar{\emph{\textbf{f}}}$ could be formulated as 
	\begin{equation}
		\begin{aligned}
			\bar{\emph{\textbf{f}}} &=G\emph{\textbf{f}}= \sum_{i=1}^{n}h(\lambda_{i})c_{i}\emph{\textbf{u}}_{i}= UH(\Lambda)\emph{\textbf{c}}= UH(\Lambda)U^{\top}\emph{\textbf{f}}\\&= U(I-\frac{\Lambda}{2})^{k}U^{-1}\emph{\textbf{f}}= (I-\frac{L_s}{2})^{k}\emph{\textbf{f}}.
		\end{aligned}
	\end{equation}
	We can also employ this low-pass filter on $X$ to achieve a smoothed representation $\bar{X}$, i.e.,
	\begin{equation}
		\bar{X}=(I-\frac{L_s}{2})^{k}X.
		\label{xbar}
	\end{equation}
	For $\bar{X}$, nearby nodes will have very similar feature values in each dimension. Eq.(\ref{xbar}) can be expanded as following
	$$
	\begin{array}{l}
		\overline{\boldsymbol{x}}_{i}^{(0)}=\boldsymbol{x}_{i}, \quad \overline{\boldsymbol{x}}_{i}^{(1)}=\frac{1}{2}\left(\overline{\boldsymbol{x}}_{i}^{(0)}+\sum_{j } \frac{w_{i j}}{\sqrt{d_{i} d_{j}}} \overline{\boldsymbol{x}}_{j}^{(0)}\right), \cdots, \\
		\overline{\boldsymbol{x}}_{i}^{(k)}=\frac{1}{2}\left(\overline{\boldsymbol{x}}_{i}^{(k-1)}+\sum_{j } \frac{w_{i j}}{\sqrt{d_{i} d_{j}}} \overline{\boldsymbol{x}}_{j}^{(k-1)}\right).
	\end{array}
	$$
	We can see that $\bar{x}_i$, i.e., $\bar{x}_i^k$, is obtained by aggregating the features of its neighbors iteratively. Notably, the $j$-th point won't contribute anything to $\bar{x}_i$ if they are not connected, i.e., $w_{ij}=0$. Hence, it incorporates long-distance data relations, which would be beneficial for downstream tasks.  
	\subsection{Multi-view Subspace Clustering}
	It can be known from the above discussion that numerous subspace clustering studies have been proposed. Let $X\in\mathbb{R}^{n\times m}$ be a set of $n$ data points, and then subspace clustering aims to identify the subspaces by expressing each sample as a linear combination of other samples.
	The math model can be simplified as
	\begin{equation}
		\min_{Z}  \|X^\top-X^\top Z\|_F^2+\alpha R(Z),
		\label{FLSR}
	\end{equation}
	where $\alpha>0$ is a trade-off parameter and $R(\cdot)$ is a regularization term. The subspace information is embedded in the coefficient matrix $Z$. {The learned coefficient matrix $Z$ is often regarded as a similarity graph and then it is fed to a spectral clustering algorithm}. Following above procedures, a great number of subspace clustering methods have been proposed \cite{lu2018subspace,peng2015robust}.
	
	Recently, multi-view subspace clustering (MVSC) has achieved significant success. Generally, for multi-view data $X=[X^1; \cdots; X^i; \cdots; X^v]\in\mathcal{R}^{\sum\limits_{i=1}^v n \times m_i}$,\\ MVSC aims to solve:
	\begin{equation}
		\min_{\{Z^i\}_{i=1}^{v}} \sum_{i=1}^{v} \|{{X^i}}^{\top}-{X^i}{Z^i}\|_F^2+\alpha R(Z^i).
		\label{msmodel}
	\end{equation}
	Here, Eq. (\ref{msmodel}) provides different solutions with different forms of $R$. For example, \cite{wang2016iterative} enforces agreement between pairs of graphs; \cite{cao2015diversity} emphasizes the complementarity of different views. In the case of multiple graphs, \cite{gao2015multi} assumes that they produce the same clustering result; \cite{cao2015diversity,wang2016iterative} perform spectral clustering on averaged graph.
	
	Nevertheless, most existing subspace clustering methods often operate on the raw data without incorporating the inherent graph structure information contained in the data points. Manifold regularization is a popular way to incorporate graph information \cite{zhai2018laplacian}, but it involves an additional term. In practice, the data might not be easy to partition in the original domain. Therefore, some methods project the data into a new space \cite{patel2015latent}. Another category of methods are inspired by the success of deep learning and implement subspace clustering with the latent representation learned by auto-encoders \cite{peng2016deep,ji2017deep}. The former class of methods are shallow techniques which lack high discriminative capability while the later are deep learning approaches which involve large number of parameters and are computationally expensive.
	
	\section{Proposed Methodology}
	Finding a suitable representation is paramount for the performance of subspace clustering. Based on the cluster assumption, adjacent points are more likely belonging to the same cluster. Putting it differently, points from the same cluster should have similar feature values. Motivated by the theory of graph filtering, we apply a low-pass graph filter on the raw data to achieve a smooth representation, which in turn makes the downstream clustering task easier.

	\subsection{Smoothed Multi-View Subspace Clustering}\label{section:smvsc}
	For multi-view data, we could  apply graph filtering strategy on it. In this paper, we choose the recently proposed large-scale multi-view subspace clustering (LMVSC) \cite{kang2019large} model to demonstrate it. This technique can obtain the partitions in $O(n)$ time. Specifically, for each view $X^i$, rather than learning a $n\times n$ graph, it constructs a smaller matrix $Z^i\in \mathbb{R}^{n\times p}$, which characterizes the relations between $p$ landmarks $\bar{A^i}\in \mathbb{R}^{m_i\times p}$ and the original data $X^i\in \mathbb{R}^{n\times m_i}$. The landmarks are supposed to well represent the whole data samples, which can be obtained by K-means or random sampling. Our proposed smoothed multi-view subspace clustering (SMVSC) can be formulated as
	\begin{equation}
		\min_{\{Z^i\}_{i=1}^{v}} \sum_{i=1}^{v} \|{\bar{X^i}}^{\top}-{\bar{A^i}}(Z^i)^{\top}\|_F^2+\alpha \|Z^i\|_F^2.
		\label{mvmodel}
	\end{equation}
	Here, we employ certain graph construction method, e.g., the probabilistic neighbor method \cite{10.5555/3016100.3016174}, to build a graph for each view ${X^i}$. {Then, for the graph filtering part, we can specify the number of filtering $k$ and a smoothed representation $\bar{X^i}={X^i}(I - \frac{L_{i}}{2})^k$ for each view is obtained. For subspace clustering part, we run K-means on $\bar{X^i}$ and let $g$ cluster centers form $\bar{A^i}$. Eq. (\ref{mvmodel}) produces $Z^i$ for each view}.
	
	Afterwards, we define $\bar{Z}=[Z^1,\cdots,Z^i,\cdots,Z^v]\in\mathcal{R}^{n\times pv}$. It has been shown that the spectral embedding matrix $Q\in\mathcal{R}^{n\times g}$, i.e., consisting of the $g$ left singular vectors, can be achieved by applying singular value decomposition (SVD) on $\bar{Z}$. Eventually, K-means is implemented on $Q$ to obtain the final partitions. The complete steps for our SMVSC method is summarized in Algorithm \ref{alg:algorithm}.
	
	Note that SMVSC algorithm is iteration-free and very efficient. Specifically, the computation of $Q$ costs $\mathcal{O}(p^3v^3+2pvn)$ and the subsequent K-means consumes $\mathcal{O}(ng^2)$. Solving $Z^i$ takes $\mathcal{O}(np^3v)$. Since $p,v\ll n$, the overall complexity is linear to the sample number.
	
	\begin{algorithm}[!htbb]
		\caption{SMVSC algorithm}
		\label{alg:algorithm}
		\textbf{Input}: Multi-view data $X^1, \cdots, X^i, \cdots, X^v\in\mathcal{R}^{\sum\limits_{i=1}^v n \times m_i}$\\
		\textbf{Parameter}: filter order $k$, trade-off parameter $\alpha$,\\ anchor number {$p$}, cluster number $g$
		\begin{algorithmic}[1] 
			\STATE Build a graph for each view by the probabilistic neighbor method
			\STATE Apply $k$ times graph filter on $X^1, \cdots, X^i, \cdots, X^v$ to obtain the smooth representation
			$\bar{X^1}, \cdots, \bar{X^i}, \cdots, \bar{X^v}$
			\STATE Run K-means on $\bar{X^i}$ form landmark $\bar{A^i}\in \mathbb{R}^{m_i\times p}$, and calculate ${Z^i}\in \mathbb{R}^{n \times p}$ by Eq. (\ref{mvmodel}), which is composed of ${\bar{Z}} \in\mathcal{R}^{n\times pv}$
			\STATE Calculate $Q$ by performing SVD on ${\bar{Z}}$
			\STATE Achieve the cluster partitions by performing K-means clustering on $Q$
		\end{algorithmic}
		\textbf{Output}: $g$ partitions
	\end{algorithm}

	\section{Multi-View Experiments} \label{sec:multi-view}
	In this section, we execute several experiments on multi-view datasets to demonstrate the effectiveness of our approach. The source code is available at  \url{ https://github.com/EricliuLiang/SMVSC}.

	\subsection{Dataset}
	Several benchmark datasets, including Handwritten, Caltech-101 and Citeseer are applied. Handwritten contains images of digits 0 to 9. Caltech-101 consists of object images, two subsets of which, i.e., Caltech-7 and Caltech-20 are commonly used in the literature. Citeseer is a citation network, whose nodes represent publications. The statistics information of above datasets are shown in Table \ref{mvdata}.  
	
	\begin{table*}[!hbtp]
		\begin{center}
			\caption{Detail information of the multi-view datasets. The feature dimension is shown in parenthesis. }
			\label{mvdata} 
			\resizebox{1.0\columnwidth}{!}{
				\begin{tabular}{llll}
					\hline
					
					{View} &{Handwritten} & {Caltech-7/Caltech-20}  & {Citeseer}  \\\hline
					1& Profile Correlations (216) & Gabor(48) &  Citation Links (3312) \\
					2& Fourier Coefficients (76) & Wavelet moments (40)&  Words Presence (3703)\\
					3& Karhunen Coefficients (64)  &CENTRIST (254)& -\\
					4 &  Morphological (6) &HOG (1984)& -\\
					5& Pixel Averages (240)  & GIST (512)& - \\
					6& Zernike Moments (47)  &LBP (928)& -  \\\hline
					Data samples & 2000 & 1474/2386& 3312 \\
					Cluster number& 10 & 7/20 & 6 \\
					\hline
			\end{tabular}}
		\end{center}
	\end{table*}

	\subsection{Comparison Methods}
	To have a convincing comparison between SMVSC and existing methods, we select several recently proposed methods that report the state-of-the-art performance.
	\begin{itemize}
		\item Parameter-Free Auto-Weighted Multiple Graph Learning (AMGL) \cite{Nie2016Parameter}: it extends spectral clustering to multi-view scenario with a novel weighting mechanism to distinguish the importance of different views.  
		\item Multi-view Low-rank Sparse Subspace Clustering (MLRSSC) \cite{Brbi2017Multi}: It develops a multi-view low-rank plus sparse subspace clustering algorithm and enforces agreements between representations of the pairs of views or a common centroid.
		\item Multi-view Subspace Clustering with Intactness-aware Similarity (MSC\_IAS) \cite{wang2019multi}: it constructs the similarity in the intact space by assuming that it has maximum dependence with its corresponding intact space, which is measured by the Hilbert–Schmidt Independence Criterion (HSIC).
		\item Large-scale Multi-View Subspace Clustering (LMVSC) \cite{kang2019large}: it addresses the scalability issue of multi-view subspace clustering method by employing anchor strategy.
	\end{itemize}
	\subsection{Experimental Setup}
	As introduced in section \ref{section:smvsc}, we first apply probabilistic neighbor method \cite{10.5555/3016100.3016174} to obtain the low-pass filter and then achieve the smooth representation for each view. {For the Handwritten data, the anchor number $p$ is searched from the range $[g,20,30,40,50,60,70,80,90,100]$ and $\alpha$ is searched in $[0.01, 0.1 ,1, 10, 100, 1000]$; the Caltech-7,  $p$ is searched from the range $[g,50,100,150,200,250,300]$ and $\alpha$ is searched in $[0.001, 0.01, 0.1, 1,10,100, 1000 ]$; the Caltech-20,  $p$ is searched from the range $[g,50,100,150,200,225,250]$ and $\alpha$ is searched in $[5, 10 ,15, 20, 25, 30, 35,\\ 40, 45, 50]$; the Citeseer,  $p$ is searched from the range $[g,20,30,40,50,60,70,80,90,\\100]$ and $\alpha$ is searched in $[1, 10 ,100, 1000,10000]$}.
	
	Clustering performance is evaluated by three commonly used metrics, including accuracy (ACC), normalized mutual information (NMI), and purity (PUR) \cite{kang2020partition}. Besides clustering performance, we also test the time consumed by these methods based on a computer equipped with a 2.6GHz Intel Xeon CPU and 64GB RAM, Matlab R2016a. 
	
		\begin{table}[htbp]
		\caption{Clustering performance on multi-view datasets($k=1$). }
		\label{mvresults1}
		\centering
		\setlength{\tabcolsep}{3.5mm}{
			\begin{tabular}{|c| c| c| c| c| c|}
				\hline
				{Data}&{Method}  & ACC & NMI & PUR & TIME(s) \\
				\hline
				\multirow{5}{*}{Handwritten}&AMGL \cite{Nie2016Parameter}& 84.60& 87.32& 87.10& 67.58\\
				&MLRSSC \cite{Brbi2017Multi}& 78.90& 74.22& 83.75& 52.44 \\
				&MSC\_IAS \cite{wang2019multi}& 79.75& 77.32& 87.55& 80.78\\
				&LMVSC \cite{kang2019large}& 91.65& 84.43& 91.65& 10.55\\
				&SMVSC&\textbf{94.30}&	\textbf{88.95}&	\textbf{94.30}&	8.58\\
				\hline
				\multirow{5}{*}{Caltech-7}&AMGL \cite{Nie2016Parameter}& 45.18& 42.43& 46.74& 20.12\\ 
				&MLRSSC \cite{Brbi2017Multi}& 37.31& 21.11& 41.45& 22.26\\ 
				&MSC\_IAS \cite{wang2019multi}& 39.76& 24.55& 44.44& 57.18\\
				&LMVSC \cite{kang2019large}& 72.66& 51.93& 75.17& 135.79\\
				&SMVSC& \textbf{73.54}&	\textbf{52.04}&	\textbf{84.87}&	236.32\\
				\hline
				\multirow{5}{*}{Caltech-20}&AMGL \cite{Nie2016Parameter}& 30.13& 40.54& 31.64& 77.63\\ 
				&MLRSSC \cite{Brbi2017Multi}& 28.21& 26.70& 30.39& 607.28 \\ 
				&MSC\_IAS \cite{wang2019multi}& 31.27& 31.38& 33.74& 93.87\\
				&LMVSC \cite{kang2019large}& 53.06& \textbf{52.71}& 58.47& 342.97\\
				&SMVSC& \textbf{56.92}&	51.90&	\textbf{64.42}&	447.58\\
				\hline
				\multirow{5}{*}{Citeseer}&AMGL \cite{Nie2016Parameter}& 16.87& 0.23& 16.87& 449.07\\ 
				&MLRSSC \cite{Brbi2017Multi}& 25.09& 02.67& 63.70& 106.10\\ 
				&MSC\_IAS \cite{wang2019multi}& 34.11& 11.53& \textbf{80.76}& 191.29\\
				&LMVSC \cite{kang2019large}& 52.26& \textbf{25.71}& 54.46& 21.33\\
				&SMVSC& \textbf{55.40}&	25.57&	57.27&	21.82\\
				\hline
			\end{tabular}
		}
	\end{table}
	
	\subsection{Results}
	In this experiment, we fix graph filter order $k=1$, and the experiment results are summarized in Table \ref{mvresults1}. As we can see, our proposed method SMVSC often achieves the best performance. In particular, SMVSC consistently enhances the accuracy of LMVSC on four datasets. In terms of NMI, our method produces better or comparable results as LMVSC. As for PUR, our proposed method outperforms others for three datasets except the last one. Note that the only difference between SMVSC and LMVSC lies in that SMVSC adopts graph filtering to preprocess the data. Both SMVSC and LMVSC often outperform other techniques by a large margin. This could be explained by their inherent drawbacks. For example, AMGL uses the inversion of loss as the weight for each view, which is too restrictive in practice; MLRSSC 
	imposes both low-rank and sparse constraints, which could lead to conflict solutions; MSC\_IAS employs a fixed weight for each view, which fails to explore the heterogeneity in views.
	
	In terms of computation time, our method is also competitive with respect to others. It can be seen that our time fluctuates a lot on different datasets. This could be explained by the fact that our complexity is closely related to the number of anchors and different numbers are used for different datasets. Similarly, LMVSC is also heavily influenced by the number of anchors. These verify the effectiveness and efficiency of SMVSC.
	
	\begin{figure*}[!hbtp]
		\centering
		\subfigure{
			\includegraphics[width=0.30\textwidth]{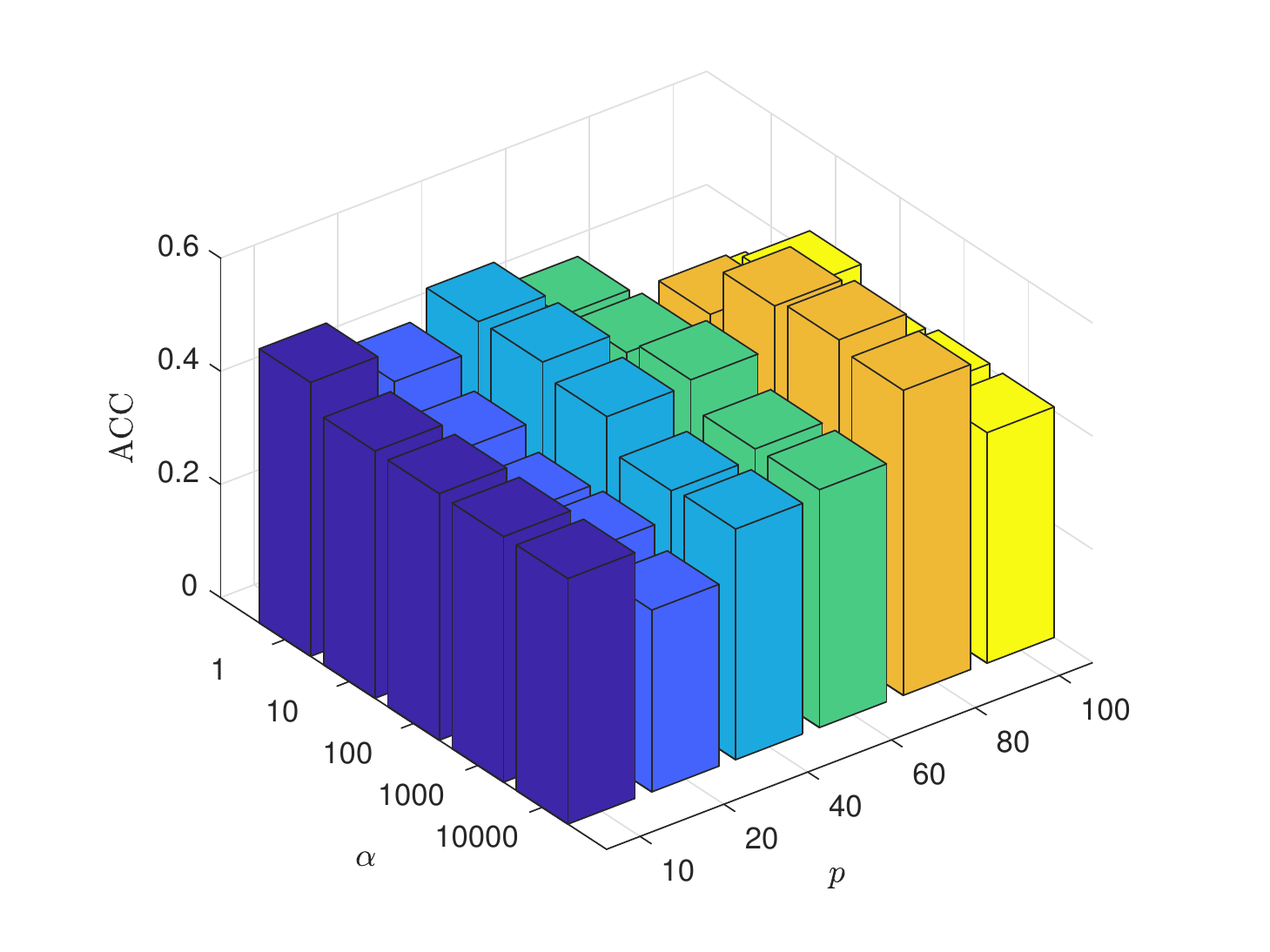}}
		\subfigure{
			\includegraphics[width=0.30\textwidth]{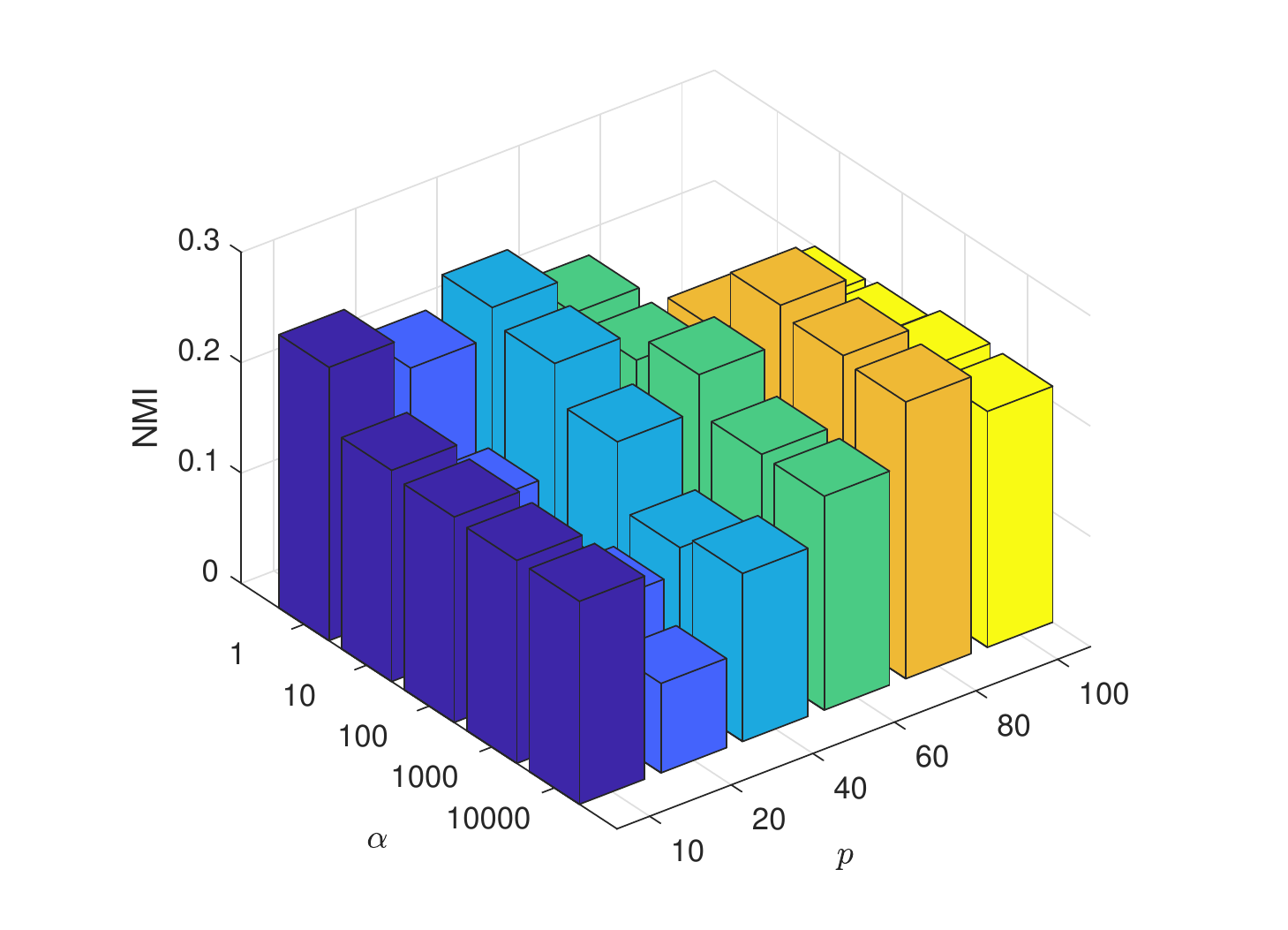}}
		\subfigure{
			\includegraphics[width=0.30\textwidth]{nmi-eps-converted-to.pdf}} 
		\caption{The influence of parameters $\alpha$ and $p$ for SMVSC on Citeseer dataset.}
		\label{Citeseer_parameter}
	\end{figure*}
	
	\begin{figure*}[!h]
		\centering
		\subfigure[$k=0$]{
			\includegraphics[width=0.3\textwidth]{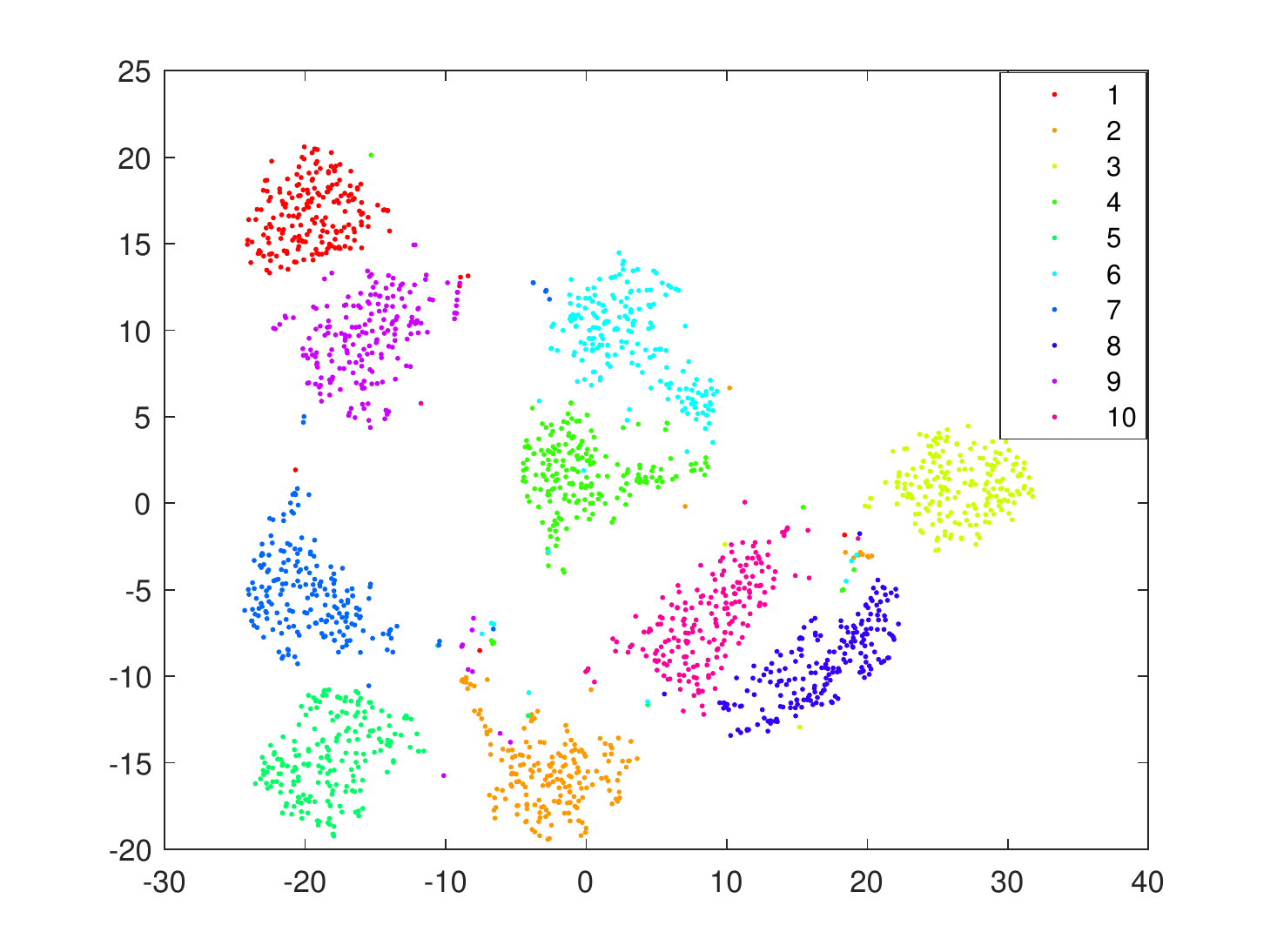}}
		\subfigure[$k=1$]{
			\includegraphics[width=0.3\textwidth]{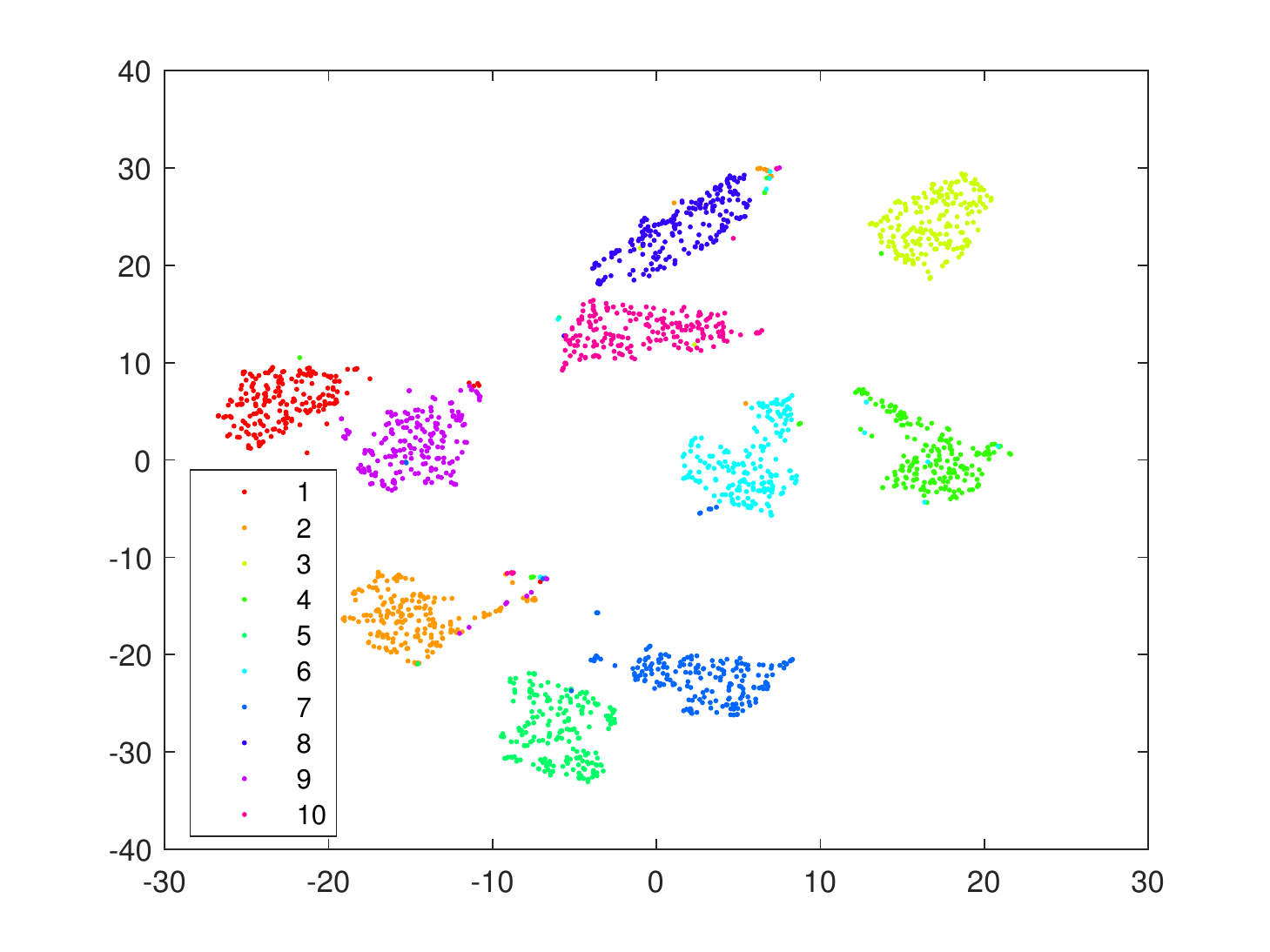}}
		\subfigure[$k=5$]{
			\includegraphics[width=0.3\textwidth]{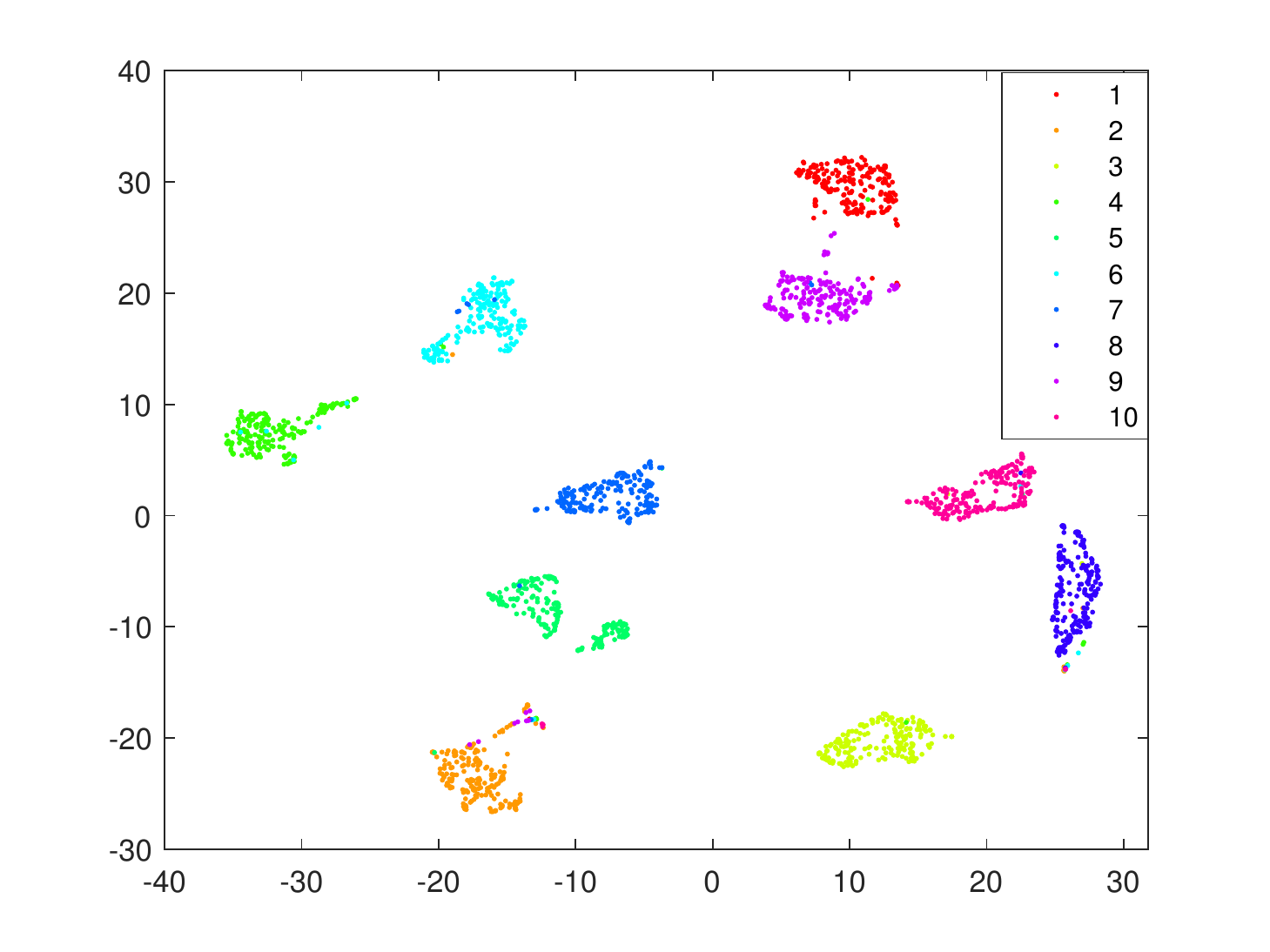}}
		\caption{The visualization of smooth representation in different graph filter order $k$.}
		\label{Handwritten_tsne}
	\end{figure*}

	\begin{table}[htbp]
		\caption{The influence of graph filter order $k$ on ACC, NMI, PUR for two datasets. }\label{diffres}
		\label{results:deep}
		\begin{tabular}{lccccccccccccc}
			
			\hline
			Data & Metric & Unfiltered & $k=1$ & $k=2$ & $k=3$ & $k=4$ & $k=5$ & $k=6$ & $k=7$ & $k=8$ & $k=9$ & $k=10$\\
			\hline
			\multirow{3}{*}{{Handwritten}}& ACC & 91.65 & 94.30 & {93.25} & {93.30} & {93.80} & {95.40} & {95.35} & \textbf{95.65} & 95.50 & 95.50 & 95.35\\
			& NMI & 84.43 & 88.95 & {87.55} & {88.14}  & {88.60} & {90.94} & {90.90} & {91.21} & 91.44 & \textbf{91.66} & 91.52\\
			& PUR & 91.65 & 94.30 & {93.25} & {93.30} & {93.80} & {95.40} & {95.35} & \textbf{95.65} & 95.50 & 95.50 & 95.35\\
			\hline
			\multirow{3}{*}{Caltech-7}& ACC & 72.66 & 73.54 & \textbf{76.79} & {72.72} & {72.11} & {73.54} & {72.04} & {72.11} & 70.96 & 72.93 & 70.01\\
			& NMI & 51.93 & 52.04 & \textbf{56.06} & {50.43}  & {53.85} & {52.42} & {53.57} & {52.57} & 49.36 & 53.99 & 52.01\\
			& PUR & 75.17 & 84.87 & {83.92} & {85.27} & {84.73} & \textbf{86.02} & {82.76} & {84.12} & 84.12 & 85.34 & 79.71\\
			\hline
		\end{tabular}
	\end{table}
	
	\subsection{Parameter Analysis}
	There are three parameters in our model, including trade-off parameter $\alpha$, anchor number $p$, graph filter order $k$. Taking Citeseer as an example, we show the influence of $\alpha$ and $p$ in Fig. \ref{Citeseer_parameter}. It can be observed that, for a fixed $\alpha$, the performance can be improved by increasing $p$ to some extent.
	Nevertheless, the performance deteriorates if $p$ has a too large value. In addition, we can see that it is easy to obtain a good performance with an appropriate $p$, and so we can achieve reasonable results by fixing $p$ to a small range and tuning $\alpha$ in practice.
	 
		Table \ref{diffres} summarizes the clustering results of Handwritten and Caltech-7 datasets under different graph filter order $k$, fixed $\alpha$ and $p$. As previously discussed, when $k$ increases, the features of adjacent nodes will be more similar. Nevertheless, if $k$ is too large, it will result in over-smoothing, i.e., the features of nodes from different clusters will be mixed and lead to indistinguishable. Therefore, a too large $k$ will negatively destroy the clustering performance. Specifically, we could observe that the clustering performance keeps increasing till $k$ equals 7 for Handwritten, and $k$ reaches 2 for Caltech-7. Moreover, the reported clustering performance on two datasets are much better than that in Table \ref{mvresults1}. From this perspective, we can also improve clustering performance in other multi-view datasets by choosing an appropriate graph filter order. On the other hand, LMVSC, which doesn't employ graph filtering, generates clustering performance 0.9165, 0.8443, 0.9165 on Handwritten, 0.7266, 0.5193, 0.7517 on Caltech-7, in terms of accuracy, NMI, purity, respectively. They are inferior to our performance.
	
	To clearly see the effect of graph filtering, we apply t-SNE on Handwritten to observe the evolution process of representation $\bar{X}$ in Fig. \ref{Handwritten_tsne}. It can be seen that the smooth representation displays a clear cluster structure when filter order increases. Furthermore, we can see that the distance between clusters become larger when $k$ increases. Hence, graph filtering could increases the separability of clusters. As a result, the grouping property of smooth representation reduce the difficult of separating the data points into disjoint subspaces.

	\section{Conclusion}
	\label{conclude}
	In this paper, we propose to perform multi-view subspace clustering in a smooth representation realized by a graph filtering technique. The proposed strategy is general enough to integrate with various multi-view subspace clustering models. In particular, through a low-pass filter, the new data representation becomes more separable and is easy to cluster. Consequently, the smooth representation can boost the clustering performance. Extensive experiments on multi-view data validate the superiority of our approach. Experimental analysis shows that graph filtering can increase the separability of clusters, which also explain why it improves clustering performance. \\

	\noindent\textbf{Acknowledgments}. This paper was in part supported by Grants from the Natural
    Science Foundation of China (Nos. U19A2059, 61806045), the National Key R\&D Program
    of China (Nos. 2018AAA0100204, 2018YFC0807500), the Sichuan Science and Techology Program
    (No. 2020YFS0057), the Fundamental Research Fund for the Central
    Universities under Project ZYGX2019Z015, the Ministry of Science and Technology of
    Sichuan Province Program (Nos. 2018GZDZX0048, 20ZDYF0343, 2018GZDZX0014, 2018GZDZX0034).

	
	

	\bibliographystyle{splncs04}
	\bibliography{mybibliography}

\end{document}